# Classroom Behavior Monitoring with YOLO An Empirical Study in Higher Education Settings

Sinh Vu Trong[1[0009−0004−8161−1686]], Dung Nguyen Manh[2[0009−0007−1299−3619]], Hieu Hoang Minh[3[0009−0008−5032−1204]], Hieu Pham Trung[4[0009−0002−8374−4912]], Thu Pham Ha[5[0009−0005−5101−1863]], and Nhu Le Hoang[6[0009−0006−4351−5866]]

[1] Banking Academy of Vietnam, Hanoi, Vietnam sinhvt@hvnh.edu.vn
[2] Banking Academy of Vietnam, Hanoi, Vietnam nguyenmanhdung0627@gmail.com
[3] Banking Academy of Vietnam, Hanoi, Vietnam 26a4041235@hvnh.edu.vn
[4] Banking Academy of Vietnam, Hanoi, Vietnam hieupt05.work@gmail.com
[5] Banking Academy of Vietnam, Hanoi, Vietnam thucan041105@gmail.com
[6] Banking Academy of Vietnam, Hanoi, Vietnam hoangnhule017@gmail.com

**Abstract.** Classroom behavior monitoring plays a vital role in evaluating student engagement and improving teaching effectiveness. Traditional observation methods remain subjective and lack scalability. This study introduces a real-world dataset of classroom videos collected at the Banking Academy of Vietnam (BAV-Classroom dataset), annotated with nine distinctive behavioral categories. State-of-the-art Computer Vision models were evaluated and compared, with YOLOv11 achieving the best performance. Experimental results indicate that students' concentration often decreases notably during the final part of lectures, highlighting challenges in sustaining engagement. Our findings demonstrate the feasibility of applying computer vision for automated classroom monitoring, providing valuable insights for academic quality management.

## 1 Introduction

In the current educational environment, classroom activities generate a diverse and abundant range of data, which differs significantly from traditional classroom settings that primarily rely on textual [2], [14] or verbal information [18]. Among these data types, visual data plays a pivotal role in reflecting the dynamics of teaching and learning interactions. Specifically, classroom images can offer rich visual cues about students' emotional expressions and attention levels [18].

However, the analysis of this type of data mainly relies on manual observation and subjective interpretation, which can lead to bias and limit the scalability of its application [2]. To address this challenge, many recent works have sought to automate classroom observation and assessment through image-based analysis methods. Instead of relying solely on traditional indicators such as test scores or surveys, directly analyzing classroom activities through video can provide teachers with deeper insights into students' levels of attention and engagement [3].

Computer Vision (CV) has emerged as a prominent approach to this problem, allowing systems to collect and extract information from image and video data to recognize faces and detect student behaviors, which opens up opportunities for applications in the educational domain. Integrating Computer Vision into classroom monitoring has the potential to reduce human intervention while providing objective and real-time assessments of classroom status. This not only assists teachers in monitoring classroom activities more effectively but also generates valuable data to support the improvement of teaching methodologies.

In this paper, we survey existing Computer Vision models, including *R-CNN, Fast R-CNN, Faster R-CNN, YOLO* (You Only Look Once) series, and assess their potential for application in classroom video analysis. Specifically, the contributions of the paper are summarized as follows:

1. We collect a real-world dataset from the classroom cameras, then annotated manually to ensure the correctness of the student behaviors. We also proposes *9 distinctive student behaviors* to be monitored in the classroom, which serve as the basis for object labels in the dataset.
2. Our experiments were conducted using state-of-the-art CV models, and results were evaluated on the annotated dataset BAV-Classroom dataset.
3. We provide initial recommendations for the Banking Academy of Vietnam on the potential application of a camera-based system for automated classroom quality management.

## 2 Related works

In recent years, the application of Computer Vision in education has become increasingly widespread. Notably, these methods have been applied to analyze student behaviors in the classroom, enabling the assessment of attention levels, interaction, and participation during teaching activities. Such information provides instructors with more objective insights into classroom dynamics, thereby allowing them to adapt their lectures according to the specific needs of each class.

A key strength of computer vision in education is its ability to provide realtime feedback on student engagement levels. For instance, a study in online learning environments [10] applied computer vision techniques to categorize students based on engagement levels, demonstrating the technology’s effectiveness in detecting varying degrees of participation.

Furthermore, Computer Vision applications extend beyond simple detection by enabling adaptive learning experiences. By leveraging engagement data, instructors and student affairs departments can refine teaching strategies, enhance student management, and better address individual learner needs, ultimately fostering more personalized learning environments [17]. Hariharan’s research, for example, explored gaze tracking combined with Convolutional Neural Networks (CNNs) to analyze student behavior in online classrooms during the COVID-19 era, with a focus on face and head pose recognition [12].

Due to the limitations of traditional behavior assessment methods and observer-based ratings, scaling up classroom observation studies remains challenging. An alternative is the automatic estimation of student engagement using machine learning and computer vision. Studies employing this approach have achieved average accuracy rates of up to 85% [9], demonstrating greater objectivity and superiority over traditional methods in behavior evaluation.

Despite significant achievements in applying computer vision to assess student learning behaviors, several gaps remain that our research seeks to address:

- Lack of real-world datasets: Most existing studies rely on online classroom data, which does not accurately reflect physical classroom environments. A new, diverse, and representative dataset is required.
- Absence of automated evaluation systems: Institutions have not yet deployed AI-powered systems that utilize classroom camera data to assess student participation, with evaluations still largely conducted manually.

## 3 Data collection and labeling

| Label type | Label | Description |
|---|---|---|
| Class participation | Focusing | Looking at the board/lecturer, raising hand to speak |
| | Looking_at_ Documents | Looking at or taking notes on documents |
| | Using_PC | Using a computer/tablet for learning activities |
| Non-Class participation | Eating_Drinking | Eating or drinking during class |
| | Turning_Head | Turning head to talk |
| | Using_mobile_phone | Using a mobile phone during class |
| | Unknown_Distraction | Other unclear distraction be-haviors |
| Neutral | Lecturer | Identifying the lecturer to avoid confusion with students |
| | Not_Identify | Behaviors that are unclear or cannot be identified |

**Table 1.** Behavioral labels used in the model

To train an effective computer vision model, the research team surveyed and selected classroom videos recorded at the Banking Academy of Vietnam as the primary data source. **13 videos** were collected, totaling **11 hours and 38 minutes** across **five classrooms**. All videos were captured using a fixed camera, documenting sessions with both instructors and students. The videos were segmented into frames at a rate of *1 FPS*. After extraction, the frames were carefully filtered based on three main criteria to ensure data quality: *Lighting, Camera angle, Image quality*.

In order to construct a reliable set of behavior labels aligned with the objectives of this study, our team reviewed existing classifications from prior research. The study by Trabelsi [13], which categorizes learning behaviors into two groups: high focus (such as concentrating and raising hands) and low focus (including boredom, laughing, using phones, writing, eating, etc.) Based on this classification, we introduce our dataset with 3 groups: class participation, non-class participation and neutral, lead to 9 detail labels as shown in Table 3. We create the dataset and annotate images on Roboflow[1].

# 4 Methodology

## 4.1 Computer vision models selection

R-CNN was a pioneering framework in object detection, enhancing both speed and accuracy but facing limitations in training time, real-time performance, and high computational requirements. According to Girshick [5], Fast R-CNN addressed several drawbacks of R-CNN by processing the entire image through a CNN and integrating classification with bounding-box regression [8]. However, its reliance on Selective Search still constrained speed and complicated deployment. Faster R-CNN replaced Selective Search with a Region Proposal Network [6], achieving higher mAP and faster performance, yet it continued to fall short in real-time applications and demanded substantial GPU resources [11]. Consequently, models such as YOLO, SSD, and RetinaNet have become increasingly favored.

Finally, YOLO represents one of the most advanced architectures and is prioritized for real-time object detection [4]. Compared with other models, YOLO can detect objects rapidly and accurately by formulating detection as a regression problem and directly predicting both class labels and object locations [7]. Moreover, when YOLOv4, YOLOv3, and Faster R-CNN were trained on the same dataset, YOLOv4 achieved the best results with 72% mAP and 63% recall [1]. In summary, building on prior research, our team selected YOLO as the primary model for applying computer vision in education, due to its ability to balance accuracy and processing

[1] https://roboflow.com/

speed.

### 4.2 Model training and evaluation

Throughout its development, YOLO has undergone multiple refinements, progressively improving in performance. Wang demonstrated that YOLOv4 achieved superior detection accuracy while reducing inference time by 15.6% compared to YOLOv3 [15]. Haokang Wen [16] further showed that YOLOv5 surpassed

YOLOv4 in accuracy, exhibited stronger performance in detecting small objects, and provided greater flexibility and faster processing. Comparisons among YOLOv5, YOLOv8, and YOLOv9 highlight significant advances in architecture, efficiency, and deployment capabilities.

| Version | Precision | Recall | mAP@0.5 |
|---|---|---|---|
| YOLOv8m | 64.59 | **78.29** | 75.01 |
| YOLOv9m | 69.32 | 77.95 | 77.06 |
| YOLOv11m | **74.82** | 77.29 | **78.80** |
| YOLOv12m | 73.52 | 75.80 | 77.85 |

**Table 2.** Comparison of Precision, Recall, and mAP@0.5 across YOLO versions.

Following the evaluation of advantages and limitations across versions, the models were trained on a dataset of 310 annotated images prepared using Roboflow. Training was conducted on Google Colab with a T4 GPU, using the following parameters: epochs = 100 and image size = 640 px. To assess and compare model performance, standard evaluation metrics were applied, including Precision, Recall, and mAP@0.5. The results are summarized in Table 2:

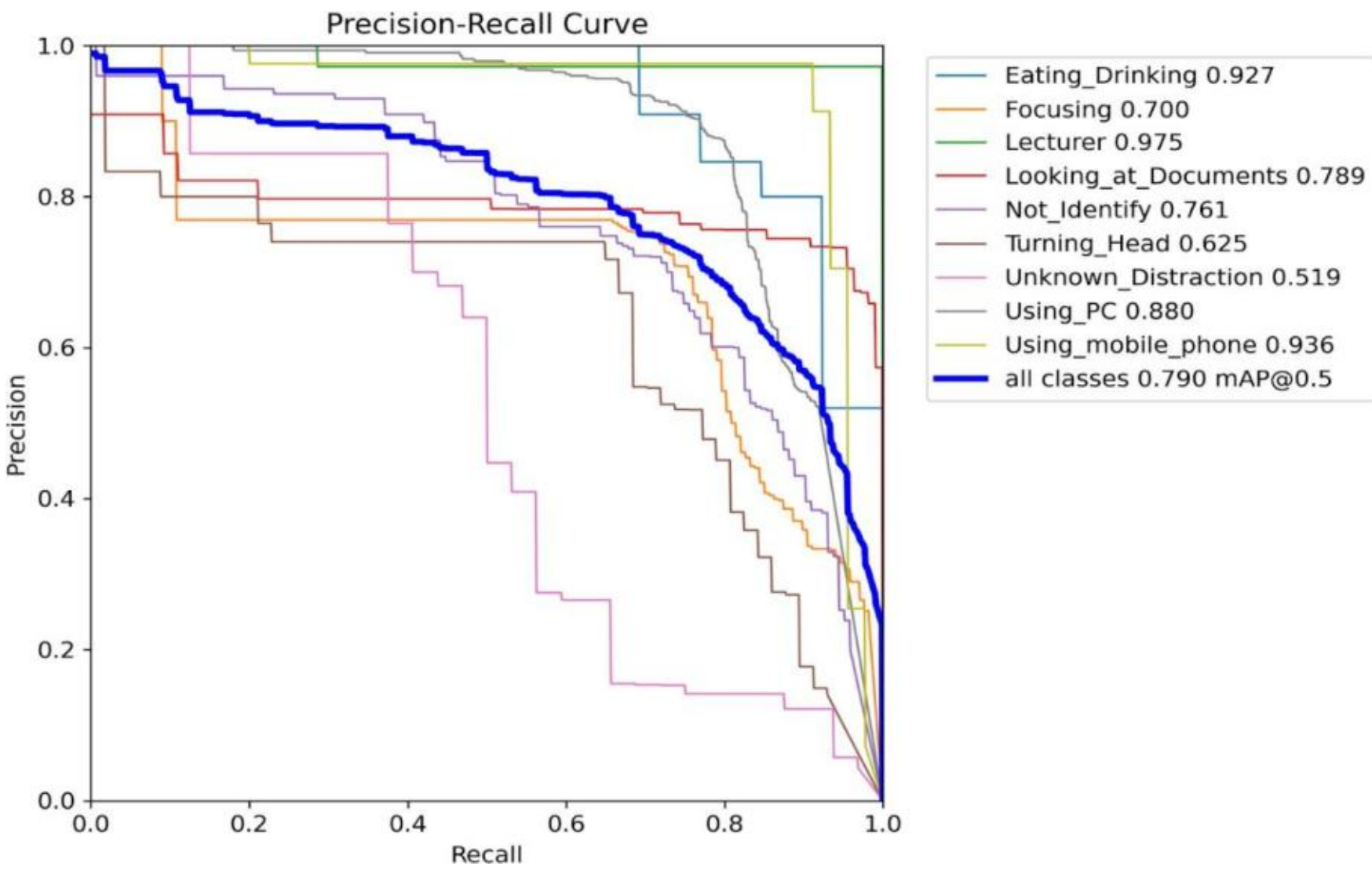

**Fig. 1.** mAP@0.5 scores of YOLO by behavior class

In terms of Precision, YOLOv11m achieved the highest score at 74.82%, followed by YOLOv12m at 73.52%, while YOLOv8m recorded the lowest at 64.59%. This indicates that YOLOv11m delivers more accurate detections with fewer false positives. Regarding Recall, YOLOv12m performed less effectively, illustrating the trade-off where higher Recall may reduce Precision. The overall performance, reflected by mAP, confirmed YOLOv11m as the strongest model with mAP@0.5 = 78.80%, whereas YOLOv8m and YOLOv9m scored lower (75.01% and 77.06%, respectively). Overall, YOLOv11m provides the best balance between Precision and Recall, demonstrating high detection efficiency, and was therefore selected for implementation in this research project.

After training YOLOv11 on the prepared dataset, the model achieved mAP = 74.82%, Precision = 77.29%, and Recall = 78.80%, indicating effective performance in behavior detection. The mAP score reflects strong overall accuracy; both Precision and Recall are stable, with Precision slightly lower than Recall but with only a minor gap, suggesting that the model is cautious in its predictions while minimizing missed detections.

To further evaluate YOLOv11, our research team relied on the Precision Recall Curve. This curve illustrates the relationship between Precision and Recall across different thresholds. According to the analysis, the average accuracy of the labels reached 0.790 mAP@0.5 , indicating that the model performs effectively in recognizing behaviors directly from input images. Examining the label-specific results, several behaviors achieved notably high performance, such as Lecturer (0.975), Using_mobile_phone (0.936), Eating_Drinking (0.927), and Using_PC (0.880). These findings demonstrate the model's strong ability to accurately detect these particular activities. Although other labels exhibited lower performance, the team is confident that with a more diverse dataset, these metrics can be further improved. Based on these outcomes, we conclude that YOLOv11 shows significant potential and is well-suited for real-world applications in assessing student engagement during classroom sessions.

## 5 Research findings

The model was tested on a 1-hour 7-minute video, with an additional IoU-based filtering algorithm combined with confidence scores to handle overlapping bounding boxes in the output. The results were generated at 25 fps and aggregated into a CSV file containing the timestamps and the frequency of detected behaviors. After exporting the results to CSV, we performed statistical analysis and obtained the following insights:

The results indicate that the most prevalent classroom behavior was Using_PC (28.19%), followed by Looking_at_Documents and Focusing, which reflect active learning participation. However, distracting or unidentified behaviors accounted for more than 50%, suggesting that students' overall level of concentration remained limited. Moreover, the high standard deviation across behaviors reveals substantial fluctuations in attention throughout the session.

Subsequently, the experimental results were evaluated through a heatmap visualization.

| Behavior name | Frequency | Average | Behavioral occurrence rate (%) | Standard Deviation |
|---|---|---|---|---|
| **Eating_Drinking** | 1673 | 0.016638 | 0.08 | 0.128765 |
| **Focusing** | 120682 | 1.200207 | 5.46 | 1.154063 |
| **Lecturer** | 45051 | 0.448041 | 2.04 | 0.552849 |
| **Looking_at_ Documents** | 137567 | 1.368132 | 6.22 | 1.019391 |
| **Not_Identify** | 405749 | 4.035256 | 18.35 | 1.704238 |
| **Turning_Head** | 268680 | 2.672077 | 12.15 | 1.612817 |
| **Unknown_ Distraction** | 412554 | 4.102933 | 18.66 | 1.828985 |
| **Using_PC** | 623359 | 6.199431 | 28.19 | 2.115228 |
| **Using_mobile_phone** | 196052 | 1.949777 | 8.87 | 1.133568 |

**Table 3.** Statistical results of behaviors with frequency, average rate, and standard deviation

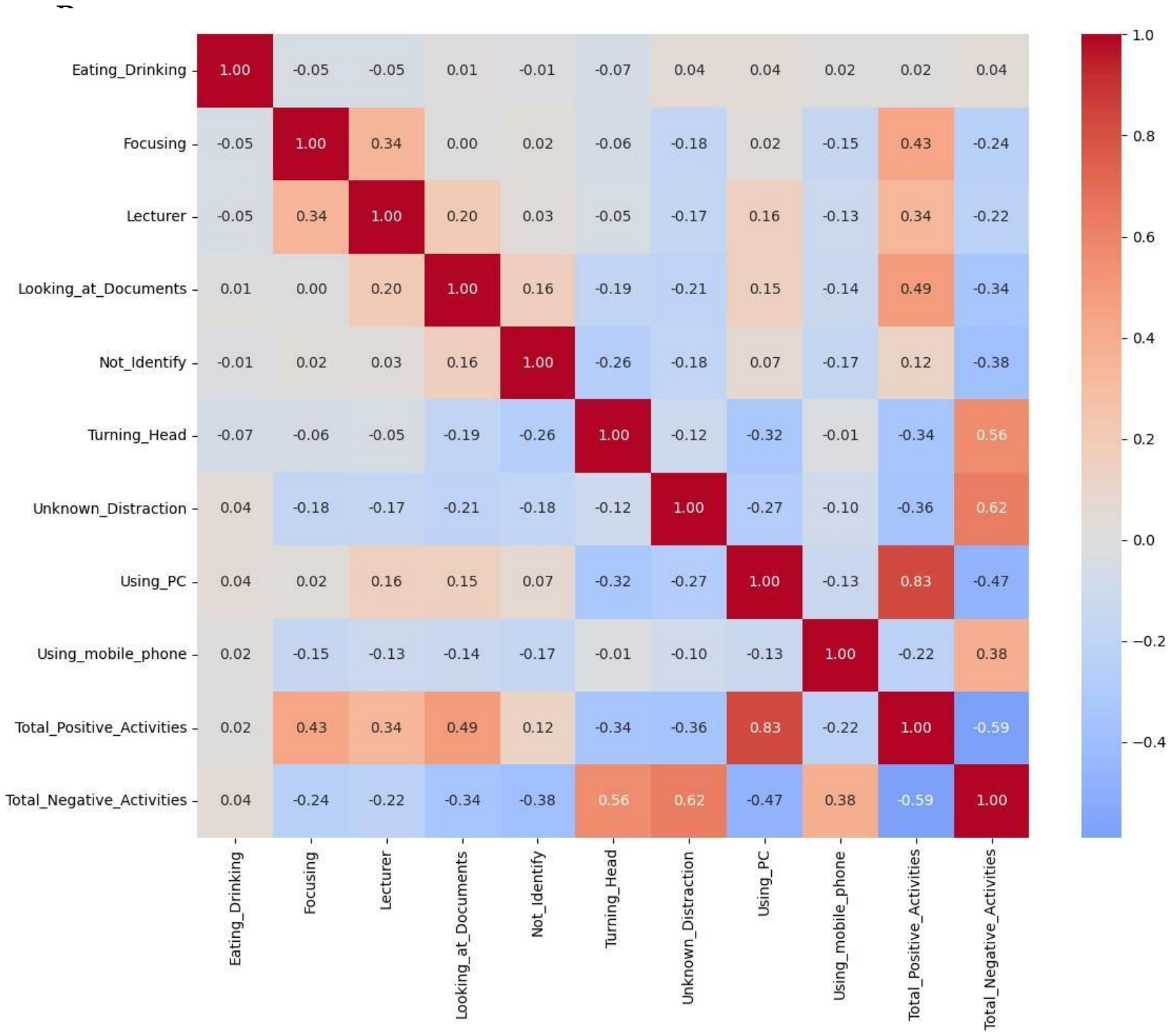

**Fig. 2.** Pearson Correlation Heatmap of Behaviors

Pearson correlation analysis (Figure 2) showed that *Using_PC* was strongly positively associated with active behaviors (r = 0.83), while *Unknown_Distraction* and *Turning_Head* were linked to negative behaviors (r = 0.62; 0.56). *Focusing*, *Looking_at_Documents*, and Lecturer negatively correlated with negative behaviors, highlighting their role in reducing distractions. Based on these findings, measures should be implemented to encourage active learning participation and reduce distractions in the classroom.

Finally, we visualized behavioral categories over time in Figure 3.

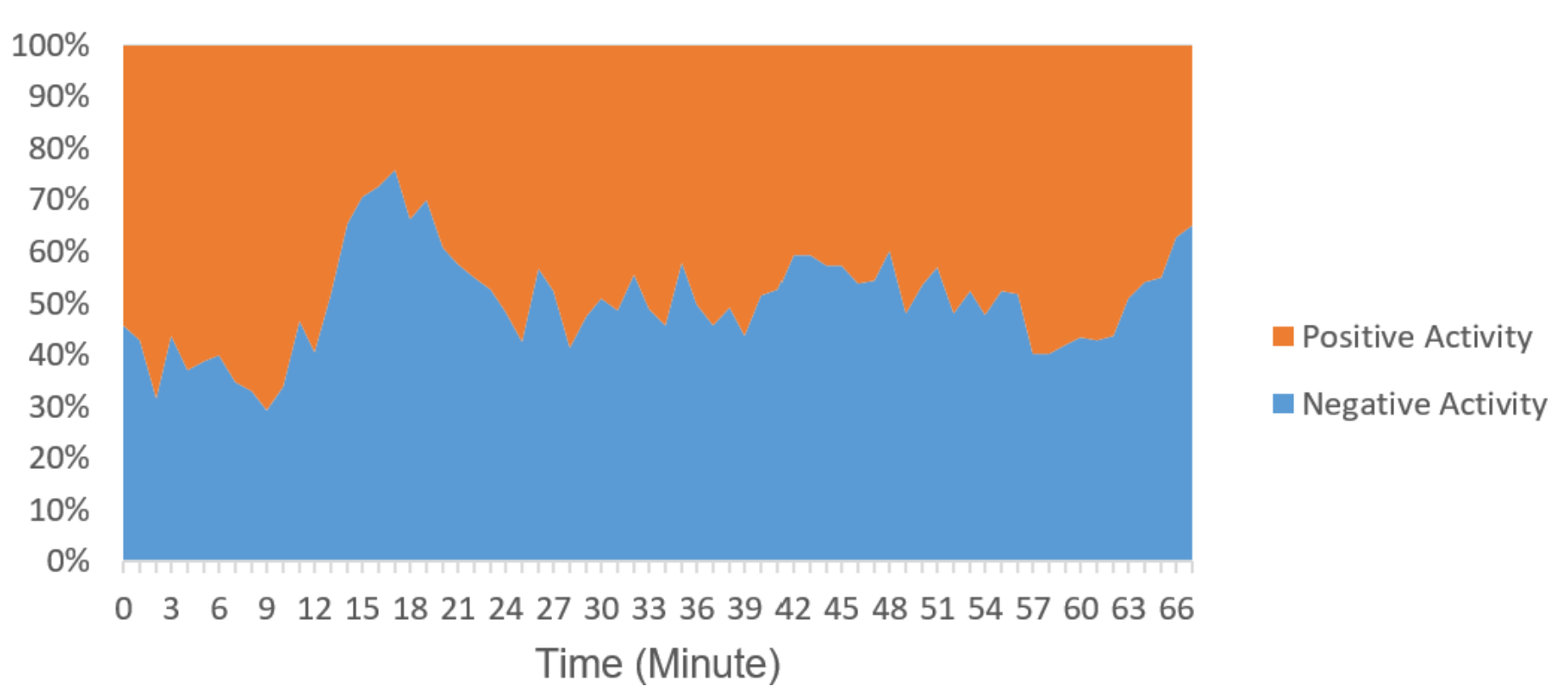

**Fig. 3.** Behavioral trends over time

The results indicate that students struggle to maintain focus over extended periods, which may be influenced by external factors or individual tendencies. Moreover, the periods immediately following the first 10 minutes and the final 10 minutes of the session were identified as the times when students are most likely to lose concentration, reflecting a common classroom behavior pattern. Based on these analyses, academic quality management departments can gain a comprehensive overview and develop strategies to support students in enhancing their knowledge retention and increasing engagement in classroom activities.

## 6 Conclusions and Future works

This study developed a computer vision system to recognize students' learning behaviors, achieving promising results with YOLOv11 (Precision 74.82%, Recall 77.29%, mAP@0.5 78.80%) on a real-world dataset collected and manually annotated from the Banking Academy of Vietnam (BAV-Classroom dataset). The model effectively identified several common behaviors, and experimental results revealed a tendency for students' attention to decline toward the end of class. These findings can be utilized to assess concentration levels, compare teaching effectiveness across sessions, and support student affairs offices in enhancing learning outcomes. Nevertheless, certain limitations remain, including the restricted training dataset, difficulties in recognizing specific behaviors, and the lack of real-time processing capability. Issues of privacy, security, and large-scale applicability also require further investigation.

In future work, the research team proposes expanding and diversifying the dataset to improve model accuracy. The system should be further optimized and integrated with advanced tracking algorithms, such as Deep SORT or transformerbased approaches, to enhance stability and enable real-time performance. For deployment, multiple cameras may be installed and combined with additional deep learning models, while privacy concerns can be addressed through clear notifications and automated data deletion mechanisms. Finally, the system holds potential for extension to diverse educational settings and examination environments to strengthen monitoring effectiveness.